\documentclass[letterpaper, 10 pt, conference]{ieeeconf}

\IEEEoverridecommandlockouts
\usepackage{amsmath,amsfonts}
\allowdisplaybreaks
\usepackage{url}
\usepackage{graphicx}
\usepackage{cite}
\usepackage{subcaption}
\usepackage{bm}
\usepackage[hang,flushmargin]{footmisc}
\usepackage{xcolor}
\usepackage[hidelinks]{hyperref}
\usepackage{eso-pic}

\title{\LARGE \bf DNPush: Distributed Nonparallel Pushing Force Control of Multi-Robot Systems for Object Transportation}

\author{
Duy Anh Nguyen$^{1}$, Pham Duy Hung$^{2}$,
and Trung Dung Ngo$^{1*}$%
\thanks{We acknowledge the support of the Natural Sciences and Engineering Council of Canada, Department of National Defence: NSERC-RGPIN-2023-05961, DGDND-2023-05961, NSERC-CREATE 528099-2019, and NSERC-Alliance ALLRP 593651-2024.}
\thanks{$^{1}$Duy Anh Nguyen and Trung Dung Ngo are with the More-Than-One Robotics Laboratory, Faculty of Sustainable Design Engineering, University of Prince Edward Island, Charlottetown, Canada.}
\thanks{$^{2}$Pham Duy Hung is with the Faculty of Electronics and Telecommunications, VNU University of Engineering and Technology, Hanoi, Vietnam.}
\thanks{$^{*}$Corresponding author. Email: {\tt\small tngo@upei.ca}}
}

\begin{document}

\maketitle

\AddToShipoutPictureFG*{%
  \AtPageLowerLeft{%
    \raisebox{4mm}{%
      \makebox[\paperwidth][c]{%
        \parbox{0.92\paperwidth}{%
          \centering
          \fontsize{6.5}{7.5}\selectfont
          This work has been submitted to the IEEE for possible publication.
          Copyright may be transferred without notice, after which this version may no longer be accessible.
        }%
      }%
    }%
  }%
}

\thispagestyle{empty}
\pagestyle{empty}
\begin{abstract}
This letter proposes DNPush, a Distributed Nonparallel Pushing force controller for cooperative object transport without inter-robot communication. Each robot uses local force measurements, a preassigned force weight, and a locally expressed desired transport direction, without requiring an object model, contact locations, a CoM estimate, or moment arm magnitudes online. DNPush regulates nonparallel force magnitudes from local force direction misalignments through a reciprocal-sine law. Under the stated contact conditions, Lyapunov analysis with LaSalle's invariance principle proves convergence to a unique rotational equilibrium, where perpendicular force contributions cancel and the net torque from normal forces is zero. Object velocity converges to the desired transport direction, and suitable contact regions exist for convex polygons. Simulations isolate the reciprocal-sine scaling mechanism, five-robot coordination, and active rotational damping. Hardware experiments with soft tactile differential-drive robots achieve a velocity direction error of $0.61^\circ\pm0.79^\circ$ over five trials with a fixed transport direction and demonstrate object transport through waypoints with three robots.

\end{abstract}

\textbf{\textit{Index Terms} -- Cooperative object transport, distributed control, multi-robot systems, tactile force control}
\section{Introduction}

Cooperative \emph{pushing} mobile robots move objects through direct body contact, unlike \emph{grasping}, which relies on mechanical attachment, or \emph{caging}, which maintains geometric enclosure~\cite{tuci2018review}. Each robot pushes the object at a contact location, which determines the direction of the applied force and its rotational effect about the object CoM~\cite{Mason,lynch1996stable}. 

For a given set of contact locations, the pushing forces must be coordinated to achieve the desired object motion. A centralized controller can allocate the force magnitudes using global information about the object geometry and pushing directions. However, global information may be limited, and reliable inter-robot communication cannot always be assumed~\cite{farivarnejad2022ar}. Distributed control reduces this dependence by allowing each robot to regulate its own pushing force using local information. The challenge is to design local control laws so that the resulting pushing forces collectively produce the desired object motion~\cite{farivarnejad2022ar,tuci2018review}. A previous distributed force controller~\cite{SII} considered parallel pushing on a common object edge. It used the local force direction misalignment $\Delta_i$ to rotate the object until the common pushing direction aligned with the desired transport direction, without inter-robot communication.

\begin{figure}[t!]
\centerline{\includegraphics[width=0.9\linewidth]{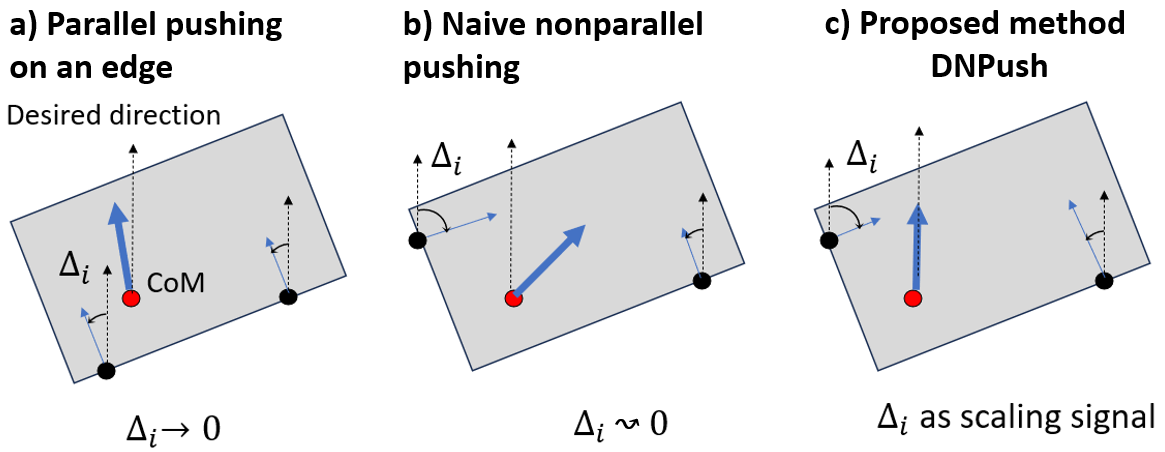}}
\caption{Core idea of DNPush. (a) For parallel pushing on a common object edge, reducing $|\Delta_i|$ aligns the common pushing direction with the desired transport direction. (b) For nonparallel contacts, the same local alignment principle does not guarantee alignment of the resultant pushing force. (c) DNPush instead uses $\Delta_i$ to scale each pushing force, coordinating the force magnitudes while stabilizing object rotation without inter-robot communication.}
\label{fig_rb}
\vspace{-5mm}
\end{figure}

This method is limited to a common pushing direction and does not directly extend to nonparallel pushing. Nonparallel contacts remove the common straight edge requirement and allow robots to push on different or curved boundary regions with different rotational effects about the object CoM. However, reducing each local force direction misalignment no longer guarantees that the resultant pushing force aligns with the desired transport direction. DNPush instead uses $\Delta_i$ to scale each pushing force magnitude, allowing the nonparallel forces to coordinate their resultant while stabilizing object rotation without inter-robot communication, as summarized in Fig.~\ref{fig_rb}.

The main contribution of this work is DNPush, a distributed controller for nonparallel pushing that achieves coordinated force allocation without inter-robot communication or moment arm magnitudes in the online controller. At equilibrium, the resulting force allocation balances both the perpendicular force contributions and the net torque from the normal forces. Under the stated contact conditions, we establish the existence and uniqueness of this equilibrium and use a Lyapunov analysis with LaSalle's invariance principle to prove asymptotic rotational convergence, followed by alignment of the normal force resultant and convergence of the transport direction. Simulations isolate the reciprocal-sine scaling mechanism in a two-robot comparison, demonstrate coordinated force allocation with five robots, and characterize the role of rotational damping. Hardware experiments validate the controller on tactile-based mobile robots.

\section{Related Work}

Cooperative object transport using multi-robot systems is commonly grouped by transport strategies, categorized as caging, grasping, and pushing ~\cite{tuci2018review}. Caging methods surround the object and maintain geometric containment during transport~\cite{wangkumar2002closure,wan2020caging}. Caging requires the robots to maintain a coordinated enclosing configuration around the object throughout transport.  Grasping methods physically attach the robots to the object, allowing them to transmit forces and torques in multiple directions. Several distributed controllers exploit these fixed connections to regulate shared object motion under limited inter-robot communication or uncertain object parameters ~\cite{wangschwager2016forceants,culbertson2021decentralized}. This bidirectional interaction supports coordinated translation and rotation, but requires secure mechanical attachments and suitable grasp points.

Using pushing forces for object transport avoids the above-mentioned requirements, but each robot is restricted to unilateral contact forces. Classical manipulation research established the mechanics and controllability of planar pushing ~\cite{Mason,lynch1996stable}. Earlier behavior- and occlusion-based methods used simple local rules to coordinate cooperative pushing without explicit communication ~\cite{chen2015occlusion}, but did not directly regulate the magnitudes of multiple nonparallel contact forces. Recent cooperative pushing studies have increasingly considered contact configuration and pushing planning. Contact locations have been optimized to improve the achievable object wrench under frictional constraints ~\cite{hung2025optimizing}. In ~\cite{tang2025pushingbots}, hybrid optimization methods have also been developed to generate and switch among their associated pushing forces for multi-robot transport in complex environments. Force- and tactile feedback methods have been developed for robust pushing under uncertain object properties, and contact changes ~\cite{toronto,Ozdamar2024_RAL} while tactile SLAM methods estimate object shape and pose online during planar pushing ~\cite{Tactile_Slame}. Most of these works address contact selection, pushing mode planning, globally coordinated force allocation, or reactive manipulation under uncertain interactions. DNPush instead considers a fixed nonparallel contact configuration and addresses how several robots coordinate their pushing force magnitudes using only local tactile feedback and no inter-robot communication.

The most related work is the previous study of Distributed Cascade Force Control (DCFC) for soft tactile multi-robot object transport~\cite{SII}. DCFC used local tactile force regulation and established rotational convergence without inter-robot communication. However, it required parallel pushing forces applied at contacts on a common straight object edge. This requirement restricts its application to objects with a sufficiently long straight boundary segment that accommodates multiple robots. DNPush overcomes this limitation by allowing \emph{nonparallel} boundary contacts of pushing forces on the transported object and changing the role of force direction misalignment. Instead of reducing the misalignment to zero, DNPush uses it to set the local force magnitude, balancing the force components perpendicular to the desired direction, and creating a restoring torque under the ordered opposite moment condition.

\section{System Modeling and Problem Formulation}
\subsection{Robot Pushing Forces and Object Dynamics}

\begin{figure}[t]
\centering
\includegraphics[width=0.8\linewidth]{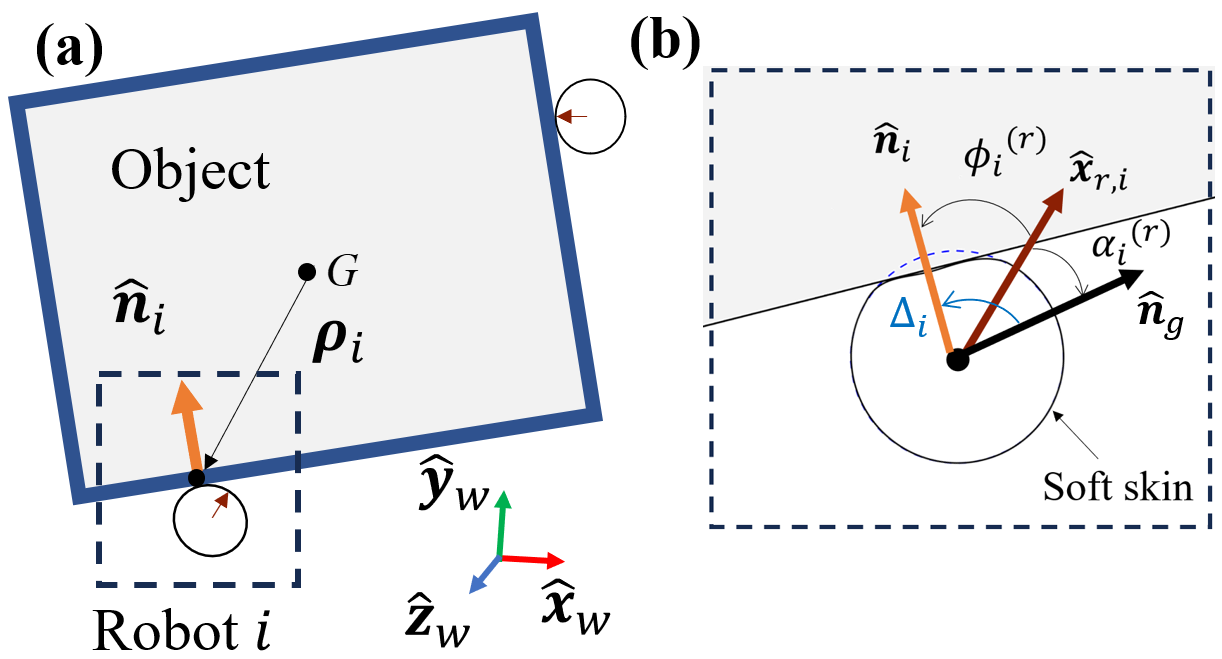}
\caption{Geometric and sensing notation for DNPush.
(a) Contact geometry.
(b) Local sensing and angular relations.}
\label{fig_notation}
\vspace{-3mm}
\end{figure}

We consider a planar rigid object pushed by $n$ differential-drive robots on a frictional surface. Let $\bm v_G\in\mathbb R^2$ denote the center-of-mass (CoM) velocity and $\omega\in\mathbb R$ denote the angular velocity about $\widehat{\bm z}$. Robot $i$ maintains a unilateral boundary contact at position $\bm\rho_i$ relative to the CoM, as shown in Fig.~\ref{fig_notation}(a).

At contact $i$, let $\widehat{\bm n}_i$ denote the local inward contact normal and let $\widehat{\bm t}_i=R(\pi/2)\widehat{\bm n}_i$ denote the local tangential direction. The physical contact force applied by robot $i$ to the object is
\begin{equation}
\bm f_i = F_{n,i}\widehat{\bm n}_i + F_{t,i}\widehat{\bm t}_i, \qquad |F_{t,i}| \le \mu_i F_{n,i},
\end{equation}
where $F_{n,i}\ge0$ is the \emph{measurable} normal contact force, $F_{t,i}$ is the \emph{unmeasured} tangential contact force, and $\mu_i$ is the local contact friction coefficient.

The planar Newton-Euler dynamics of the object are
\begin{equation}
\begin{aligned}
m\dot{\bm v}_G &= \sum_{i=1}^{n}\bm f_i+\bm F_f,\\
I_{\mathrm{cm}}\dot{\omega} &= \sum_{i=1}^{n}F_{n,i}d_{n,i} +\Delta\tau_t+M_f.
\end{aligned}
\end{equation}
where
\vspace{-0.5cm}
\begin{equation}
d_{n,i}:=(\bm\rho_i\times\widehat{\bm n}_i)\cdot\widehat{\bm z}, \qquad \Delta\tau_t:=\sum_{i=1}^{n}F_{t,i}(\bm\rho_i\times\widehat{\bm t}_i)\cdot\widehat{\bm z}.
\end{equation}
Here, $d_{n,i}$ is the signed moment arm of the normal force, and $\Delta\tau_t$ is the net torque produced by the tangential contact forces. The surface friction is modeled using smooth Coulomb and Viscous components as follows:
\begin{equation}
\begin{aligned}
\bm F_f&=-c_{0,t}\tanh\!\left(\frac{\|\bm v_G\|}{v_c}\right)\frac{\bm v_G}{\|\bm v_G\|}-c_{1,t}\bm v_G,\\
M_f&=-c_{0,r}\tanh\!\left(\frac{\omega}{\omega_c}\right)-c_{1,r}\omega.
\end{aligned}
\end{equation}
Here, $c_{0,t},c_{1,t},c_{0,r},c_{1,r}\ge0$ are the friction coefficients, $v_c,\omega_c>0$ determine the smooth transition regions, and the vector term is defined as \emph{zero} at $\bm v_G=\bm 0$.

\subsection{Local Tactile Force and Contact Direction Sensing}

Superscripts $(w)$, $(b)$, and $(r)$ denote quantities expressed in the world, object-body, and robot frames, respectively. Let $\widehat{\bm x}_{r,i}$ denote the unit vector along the $x$-axis of robot $i$'s frame.

Each robot is a differential-drive platform equipped with a $360^\circ$ soft tactile sensor~\cite{Viart}. The sensor provides the \emph{measured} normal force magnitude $F_{n,i}$ and robot-object contact direction angle $\phi_i^{(r)}$ in the robot frame, as shown in Fig.~\ref{fig_notation}(b).

Let $\widehat{\bm n}_g$ denote the shared desired transport direction, with world-frame angle $\alpha^{(w)}$. Let $\alpha_i^{(r)}$ denote the same direction expressed in robot $i$'s frame. Each robot calculates
\vspace{-1mm}
\begin{equation}
\Delta_i = \operatorname{wrap}\!\left(\phi_i^{(r)}-\alpha_i^{(r)}\right),
\end{equation}
where $\operatorname{wrap}(\cdot)$ maps an angle to $(-\pi,\pi]$. In our implementation,
$\alpha_i^{(r)}=\alpha^{(w)}-\psi_i^{(w)}$, where $\psi_i^{(w)}$ is the world-frame heading angle of robot $i$. This frame conversion uses only the shared desired direction and robot $i$'s orientation, without information from other robots. The DNPush law then uses $\Delta_i$ as its local direction signal.

\subsection{Problem Statement and Objective}

We consider $n$ robots pushing a rigid object through sustained fixed contacts on a frictional support surface with $c_{1,t}>0$. The information available to robot $i$ is limited to its measured normal force $F_{n,i}$, local force direction misalignment $\Delta_i$, preassigned force weight $f_{i0}$, and, when $k_1>0$, a local angular rate estimate $\widehat{\omega}_i$. No inter-robot communication or information about other robots' states is used during transport.

Define $\widehat{\bm n}_p:=R(\pi/2)\widehat{\bm n}_g$ as the direction perpendicular to $\widehat{\bm n}_g$. Let $\bm F_n(t):=\sum_{i=1}^{n}F_{n,i}(t)\widehat{\bm n}_i(t)$ denote the net controlled normal force. Define its components along $\widehat{\bm n}_g$ and $\widehat{\bm n}_p$ as $F_g:=\bm F_n^\top\widehat{\bm n}_g$ and $F_p:=\bm F_n^\top\widehat{\bm n}_p$, respectively. The control objectives are
\begin{equation}
\lim_{t\to\infty}\omega(t)=0,\qquad \lim_{t\to\infty}F_p(t)=0,\qquad \liminf_{t\to\infty}F_g(t)>0.
\label{eq_control_objectives}
\end{equation}

The corresponding velocity direction objective is
\begin{equation}
\lim_{t\to\infty}\widetilde{\alpha}_v(t)=0,
\end{equation}
where $v_g:=\bm v_G^\top\widehat{\bm n}_g$ and $v_p:=\bm v_G^\top\widehat{\bm n}_p$. For $\bm v_G\neq\bm 0$, the velocity direction error is defined as $\widetilde{\alpha}_v:=\operatorname{atan2}(v_p,v_g)$.

Arbitrary nonparallel contacts may not generate opposite moments or allow force balance perpendicular to the desired direction. We therefore seek a nonempty class of fixed contact sets for which the objectives in~\eqref{eq_control_objectives} can be achieved. We also seek a distributed force magnitude law that uses only the local information listed above.

Note that global contact planning, selection, and repositioning as well as contact switching are outside the scope of this work.
\section{DNPush: Contact Condition and Local Force Magnitude Coordination}

\subsection{Ordered Opposite Moment Contact Condition}

We first define a class of nonparallel contact sets that can generate moments of opposite sign while allowing force balance perpendicular to the desired transport direction. For an active contact set $\mathcal A$, excluding contacts with $d_{n,i}=0$, define
\begin{equation}
\mathcal N:=\{i\in\mathcal A\mid d_{n,i}<0\},\qquad
\mathcal P:=\{i\in\mathcal A\mid d_{n,i}>0\}.
\end{equation}

For a desired transport direction $\widehat{\bm n}_g$, the active contact set satisfies the \emph{ordered opposite moment contact condition} if both groups are nonempty and, on a consistent angular branch,
\begin{equation}
\max_{j\in\mathcal N}\Delta_j<0<
\min_{i\in\mathcal P}\Delta_i,
\qquad
\max_{i\in\mathcal P}\Delta_i-
\min_{j\in\mathcal N}\Delta_j<\pi.
\label{eq_ordered_opposite_moment_condition}
\end{equation}
The first inequality makes the misalignment negative for all contacts in $\mathcal N$ and positive for all contacts in $\mathcal P$. The second limits the angular span of all active contact normals to less than $\pi$. For fixed contacts and a fixed desired transport direction, the object orientations satisfying these inequalities define the admissible angular branch used in the convergence analysis below. The condition also implies
\begin{equation}
d_{n,i}\operatorname{sgn}(\Delta_i)>0,
\qquad
\forall i\in\mathcal A.
\label{eq_sign_compatible_compact}
\end{equation}
Section~\ref{sub_Existence of the Contact Condition} shows that such contact sets exist for convex polygonal objects.

\subsection{Cascade Tactile Control Architecture}

DNPush uses a cascade control architecture. The outer loop computes the commanded normal force $f_{i,\mathrm{cmd}}$ from the local misalignment $\Delta_i$. The inner tactile force controller, adapted from~\cite{SII}, adjusts the translational velocity of robot $i$ to regulate the measured normal force $F_{n,i}$ toward its command. A separate heading controller aligns the robot heading with the desired transport direction.

\subsection{Distributed Pushing Force Control Law}

Define $[x]_+:=\max\{0,x\}$. The distributed outer loop policy is
\begin{equation}
f_{i,\mathrm{cmd}} = \left[ f_{i0} \left(\frac{k_0}{\sin|\Delta_i|+\varepsilon} - k_1\widehat{\omega}_i\operatorname{sgn}(\Delta_i) \right) \right]_+ \label{eq_force_policy}
\end{equation}
where $f_{i0}>0$ is a fixed force weight, $k_0>0$ sets the overall force level, $k_1\ge0$ provides active rotational damping through the local angular rate estimate $\widehat{\omega}_i$, and $\varepsilon>0$ is a regularization parameter that prevents division by zero when $\sin|\Delta_i|$ is small. For fixed contacts and a fixed desired transport direction, $\dot{\Delta}_i=\omega$ on a consistent angular branch, so $\widehat{\omega}_i$ may be estimated locally from the $\Delta_i$ signal. In the convergence analysis and simulations using $k_1>0$, we assume $\widehat{\omega}_i=\omega$.

DNPush regulates only the force magnitude; the pushing direction is determined by the local contact normal.

\noindent\textbf{Assumption 1 (Group-balanced force weights).}
The force weights satisfy
\begin{equation}
\sum_{i\in\mathcal P} f_{i0}
=
\sum_{j\in\mathcal N} f_{j0}.
\label{eq_group_balance}
\end{equation}
 For the nominal reciprocal-sine term with $k_1=0$ and $\varepsilon=0$, $f_i\sin\Delta_i = k_0 f_{i0}\operatorname{sgn}(\Delta_i)$. Therefore, Assumption~1 cancels the normal-force components perpendicular to the desired transport direction.
The two groups need not contain the same number of robots. For a selected group weight $S>0$, a simple choice is
$f_{i0}=S/|\mathcal P|$ for $i\in\mathcal P$ and
$f_{j0}=S/|\mathcal N|$ for $j\in\mathcal N$.
The weights are assigned offline for the selected fixed contact set, and the active contact set and moment sign groups remain unchanged during transport.

During transport, each robot evaluates~\eqref{eq_force_policy} using only its local measurements, without inter-robot communication. Coordination emerges through the object motion: object rotation changes each robot's local $\Delta_i$, thereby changing its applied force and the resulting net torque on the object.

\section{Feasibility and Convergence Analysis}

\subsection{Existence of the Contact Condition}
\label{sub_Existence of the Contact Condition}

\begin{figure}[t]
\centering
\includegraphics[width=0.75\linewidth]{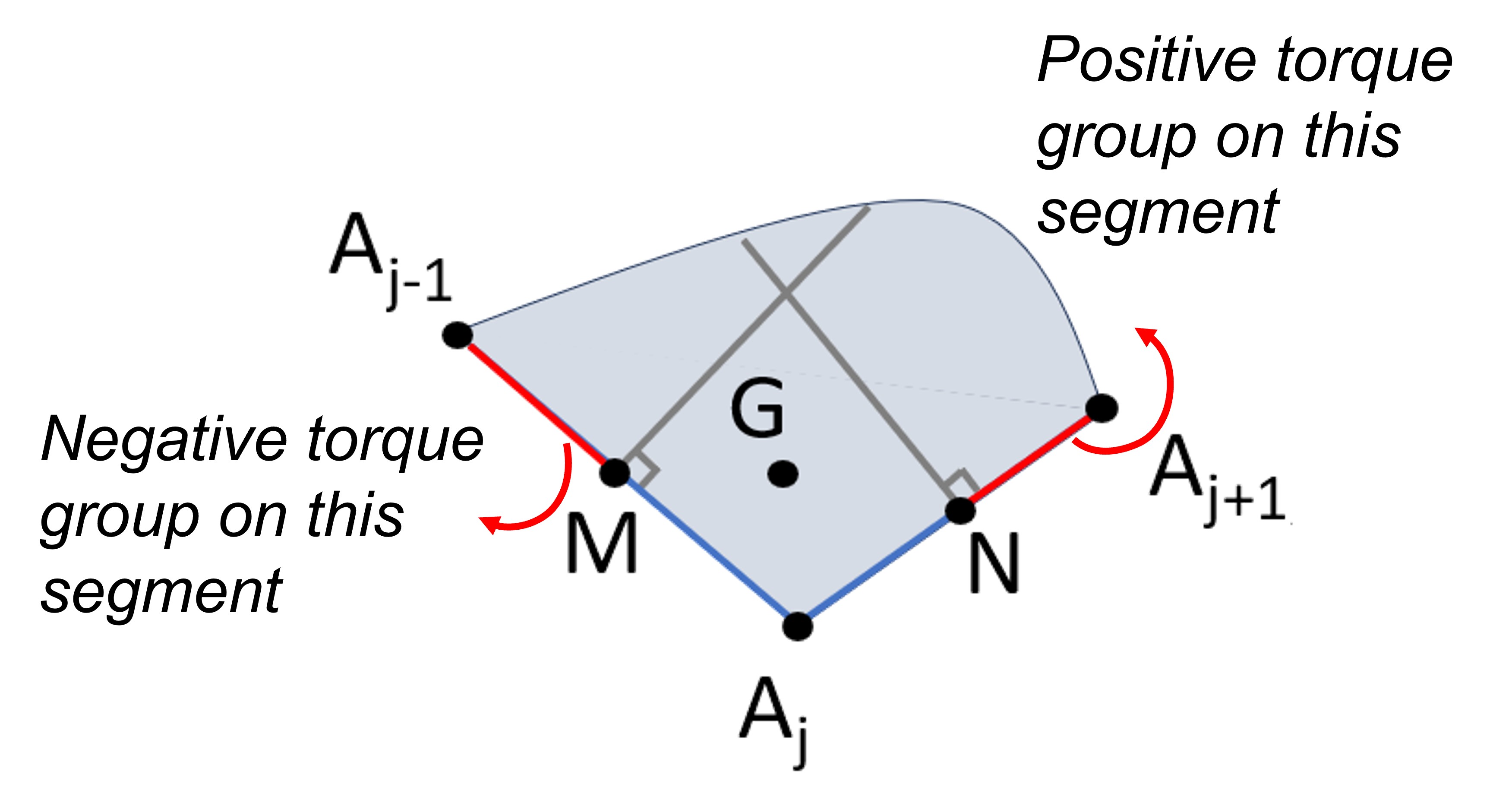}
\caption{Guaranteed negative and positive moment contact regions near the vertex of a convex polygon closest to the CoM. The inward normals of the two adjacent edges have an angular separation smaller than $\pi$.}
\label{fig_sign_compatible}
\vspace{-5mm}
\end{figure}

Fig.~\ref{fig_sign_compatible} shows the guaranteed contact regions used in the proposition below.

\noindent\textbf{Proposition 1 (Existence of an ordered opposite moment contact condition).}
Let $\mathcal O$ be a convex polygon with counterclockwise vertices $\{A_1,\ldots,A_m\}$ and no redundant collinear vertices. Assume that its CoM $G$ lies strictly inside $\mathcal O$, and let $A_j$ be a vertex closest to $G$. Every point on the open segment from $A_{j-1}$ to the midpoint of $[A_{j-1},A_j]$ has $d_n<0$. Every point on the open segment from the midpoint of $[A_j,A_{j+1}]$ to $A_{j+1}$ has $d_n>0$. The angular separation between the inward normals of these two edges is smaller than $\pi$.

\noindent\emph{Proof.}
Let $\ell^-:=\|A_j-A_{j-1}\|$ and $\ell^+:=\|A_{j+1}-A_j\|$. Define the counterclockwise unit edge tangents as $\widehat{\bm t}^{-}:=(A_j-A_{j-1})/\ell^-$ and $\widehat{\bm t}^{+}:=(A_{j+1}-A_j)/\ell^+$, with inward normals $\widehat{\bm n}^{\pm}:=R(\pi/2)\widehat{\bm t}^{\pm}$.

For a point $\bm q$ on either edge,
\begin{equation}
d_n(\bm q)=\left((\bm q-\bm G)\times\widehat{\bm n}\right)\cdot\widehat{\bm z}=(\bm q-\bm G)^\top\widehat{\bm t}.
\end{equation}

Let $M:=(A_{j-1}+A_j)/2$ and $N:=(A_j+A_{j+1})/2$. Since $A_j$ is no farther from $G$ than either adjacent vertex, $(M-G)^\top\widehat{\bm t}^{-}\le0$ and $(N-G)^\top\widehat{\bm t}^{+}\ge0$. For $0<s<\ell^-/2$ and $0<r<\ell^+/2$, define $q^-:=M-s\widehat{\bm t}^{-}$ and $q^+:=N+r\widehat{\bm t}^{+}$. Then,
\begin{equation}
\begin{aligned}
d_n(q^-) &= (M-G)^\top\widehat{\bm t}^{-}-s<0,\\
d_n(q^+) &= (N-G)^\top\widehat{\bm t}^{+}+r>0.
\end{aligned}
\end{equation}

Finally, because the polygon is convex and the two adjacent edges are not collinear, the counterclockwise angle from $\widehat{\bm t}^{-}$ to $\widehat{\bm t}^{+}$ lies in $(0,\pi)$. Rotating both tangents by $\pi/2$ preserves this angle. Hence, the counterclockwise angle from $\widehat{\bm n}^{-}$ to $\widehat{\bm n}^{+}$ also lies in $(0,\pi)$. \hfill$\square$

When the desired transport direction lies between $\widehat{\bm n}^{-}$ and $\widehat{\bm n}^{+}$, one contact from each region satisfies~\eqref{eq_ordered_opposite_moment_condition}. The subsequent analysis applies to any active contact set satisfying this condition.

\subsection{Net Restoring Torque and Equilibrium}

For the nominal convergence analysis, assume ideal inner-loop force tracking, no upper force saturation, $\varepsilon=0$, and $\widehat{\omega}_i=\omega$. Write $F_{n,i}=f_{i,\mathrm{cmd}}=:f_i$ and consider any $k_1\ge0$. Consider a fixed active contact set $\mathcal A$ satisfying~\eqref{eq_ordered_opposite_moment_condition}.

Let $\theta$ denote the counterclockwise angle from the world-frame $x$-axis to the object-fixed $x$-axis, with $\dot{\theta}=\omega$. For each fixed contact $i\in\mathcal A$, let $\phi_i^{(b)}$ denote its constant normal angle in the object frame. The corresponding world-frame normal angle is $\theta+\phi_i^{(b)}$, so the force direction misalignment is
$\Delta_i(\theta)=\operatorname{wrap}\!\left(\theta+\phi_i^{(b)}-\alpha^{(w)}\right)$.
For compactness, define $a_i:=\alpha^{(w)}-\phi_i^{(b)}$. Then
\begin{equation}
\Delta_i(\theta)=\operatorname{wrap}(\theta-a_i).
\label{eq_orientation_misalignment}
\end{equation}

For any $i\in\mathcal P$ and $j\in\mathcal N$, \eqref{eq_orientation_misalignment} gives $\Delta_i-\Delta_j=a_j-a_i$. Combining this relation with~\eqref{eq_ordered_opposite_moment_condition} gives
\begin{equation}
0<a_j-a_i<\pi,\qquad \forall i\in\mathcal P,\ \forall j\in\mathcal N.
\label{eq_ordered_active_cone}
\end{equation}

Define $\Theta:=(\theta_{\min},\theta_{\max})$, where $\theta_{\min}:=\max_{i\in\mathcal P}a_i$ and $\theta_{\max}:=\min_{j\in\mathcal N}a_j$. Equation~\eqref{eq_ordered_active_cone} gives $\theta_{\min}<\theta_{\max}$, so $\Theta$ is nonempty.

For $\theta\in\Theta$, define $x_i:=\theta-a_i$ for $i\in\mathcal P$ and $y_j:=a_j-\theta$ for $j\in\mathcal N$. Then $x_i>0$, $y_j>0$, and $x_i+y_j=a_j-a_i<\pi$, so both angles lie in $(0,\pi)$. The wrapping operation in~\eqref{eq_orientation_misalignment} is therefore inactive, giving $\Delta_i=x_i>0$ for $i\in\mathcal P$ and $\Delta_j=-y_j<0$ for $j\in\mathcal N$.

Thus, the sign compatibility in~\eqref{eq_sign_compatible_compact} holds throughout $\Theta$. Define $s_i:=\operatorname{sgn}(\Delta_i)$. Then $d_{n,i}=s_i|d_{n,i}|$, and $s_i$ remains constant on $\Theta$. Substituting the nominal assumptions into~\eqref{eq_force_policy}, the commanded normal force becomes
\begin{equation}
f_i=\left[\gamma_i(\theta)-k_1f_{i0}\omega s_i\right]_+,\qquad
\gamma_i(\theta):=\frac{k_0f_{i0}}{\sin|\theta-a_i|}.
\label{eq_projected_force_compact}
\end{equation}

The torque generated by the reciprocal-sine terms satisfies
\begin{equation}
\sum_{i\in\mathcal A}d_{n,i}\gamma_i(\theta)=k_0H(\theta),\qquad
H(\theta):=\sum_{i\in\mathcal A}\frac{f_{i0}d_{n,i}}{\sin|\theta-a_i|}.
\label{eq_restoring_H}
\end{equation}

\noindent\textbf{Lemma 1 (Unique restoring equilibrium).}
Under~\eqref{eq_ordered_active_cone}, $H$ is continuous on $\Theta$ and has a unique zero $\theta^\star\in\Theta$. Moreover,
\begin{equation}
H(\theta)>0\quad\text{for}\quad\theta<\theta^\star,\qquad
H(\theta)<0\quad\text{for}\quad\theta>\theta^\star.
\end{equation}

\noindent\emph{Proof.}
Write $H(\theta)=H_{\mathcal P}(\theta)-H_{\mathcal N}(\theta)$, where
\begin{equation}
H_{\mathcal P}(\theta):=\sum_{i\in\mathcal P}\frac{f_{i0}d_{n,i}}{\sin x_i},\qquad
H_{\mathcal N}(\theta):=\sum_{j\in\mathcal N}\frac{f_{j0}|d_{n,j}|}{\sin y_j}.
\end{equation}
Both functions are positive and continuous on $\Theta$.

Let $i_-\in\arg\max_{i\in\mathcal P}a_i$. As $\theta\to\theta_{\min}^{+}$, $x_{i_-}\to0^{+}$, while $y_j\to a_j-a_{i_-}\in(0,\pi)$ for every $j\in\mathcal N$. Hence, $H_{\mathcal P}\to+\infty$ while $H_{\mathcal N}$ remains finite, so $H(\theta)\to+\infty$ as $\theta\to\theta_{\min}^{+}$. Similarly, let $j_+\in\arg\min_{j\in\mathcal N}a_j$. As $\theta\to\theta_{\max}^{-}$, $y_{j_+}\to0^{+}$, while $x_i\to a_{j_+}-a_i\in(0,\pi)$ for every $i\in\mathcal P$. Hence, $H_{\mathcal N}\to+\infty$ while $H_{\mathcal P}$ remains finite, so $H(\theta)\to-\infty$ as $\theta\to\theta_{\max}^{-}$. Continuity therefore guarantees at least one zero in $\Theta$.

For uniqueness, let $\bar{\theta}\in\Theta$ be any zero of $H$. Then $H_{\mathcal P}(\bar{\theta})=H_{\mathcal N}(\bar{\theta})=:Q>0$. Define $p_i(\theta):=f_{i0}d_{n,i}/\sin x_i(\theta)$ and $q_j(\theta):=f_{j0}|d_{n,j}|/\sin y_j(\theta)$. Since $\mathrm dx_i/\mathrm d\theta=1$ and $\mathrm dy_j/\mathrm d\theta=-1$,
\begin{equation}
\frac{\mathrm dp_i}{\mathrm d\theta}=-p_i\cot x_i,\qquad
\frac{\mathrm dq_j}{\mathrm d\theta}=q_j\cot y_j.
\end{equation}

Evaluating at $\theta=\bar{\theta}$ and suppressing the argument $\bar{\theta}$ for compactness, we have
$\sum_{i\in\mathcal P}p_i=\sum_{j\in\mathcal N}q_j=Q$. Therefore,
\begin{align}
H'(\bar{\theta})
&=-\sum_{i\in\mathcal P}p_i\cot x_i-\sum_{j\in\mathcal N}q_j\cot y_j \nonumber\\
&=-\frac{1}{Q}\sum_{i\in\mathcal P}\sum_{j\in\mathcal N}
p_iq_j\left(\cot x_i+\cot y_j\right)<0,
\end{align}
where, for every $i\in\mathcal P$ and $j\in\mathcal N$,
\begin{equation}
\cot x_i+\cot y_j
=\frac{\sin(x_i+y_j)}{\sin x_i\sin y_j}
=\frac{\sin(a_j-a_i)}{\sin x_i\sin y_j}>0.
\label{eq_cotangent_pair_inequality}
\end{equation}

Since $H'(\bar{\theta})<0$ at every zero $\bar{\theta}\in\Theta$, $H$ can cross zero only from positive to negative and therefore has at most one zero. Together with the boundary limits, $H$ has a unique zero $\theta^\star$, with $H(\theta)>0$ for $\theta<\theta^\star$ and $H(\theta)<0$ for $\theta>\theta^\star$.
\hfill$\square$

\subsection{Asymptotic Rotational Stabilization}

For the case $\Delta\tau_t=0$, the rotational dynamics are
\begin{equation}
I_{\mathrm{cm}}\dot{\omega}
=\sum_{i\in\mathcal A}d_{n,i}f_i
-c_{0,r}\tanh\!\left(\frac{\omega}{\omega_c}\right)
-c_{1,r}\omega.
\label{eq_rotational_dynamics}
\end{equation}

\noindent\textbf{Theorem 1 (Asymptotic rotational convergence).}
Suppose that the active contacts remain fixed and satisfy~\eqref{eq_ordered_opposite_moment_condition}, and assume either $c_{0,r}+c_{1,r}>0$ or $k_1>0$.
Under~\eqref{eq_rotational_dynamics} and~\eqref{eq_projected_force_compact}, every initial condition $\theta(0)\in\Theta$ and $\omega(0)\in\mathbb R$ satisfies
\begin{equation}
\lim_{t\to\infty}\theta(t)=\theta^\star,\qquad
\lim_{t\to\infty}\omega(t)=0.
\end{equation}

\noindent\emph{Proof.}
Define the restoring potential
\begin{equation}
U(\theta):=-k_0\int_{\theta^\star}^{\theta}H(s)\,\mathrm ds.
\end{equation}
By Lemma~1, $U(\theta^\star)=0$ and $U(\theta)>0$ for $\theta\neq\theta^\star$. The reciprocal-sine singularities at the boundaries further imply
\begin{equation}
U(\theta)\to+\infty
\quad\text{as}\quad
\theta\to\theta_{\min}^{+}
\ \text{or}\
\theta\to\theta_{\max}^{-}.
\label{eq_potential_barrier}
\end{equation}

Consider the Lyapunov function
\begin{equation}
V(\theta,\omega):=\frac{1}{2}I_{\mathrm{cm}}\omega^2+U(\theta).
\end{equation}

Using $\dot{\theta}=\omega$, \eqref{eq_rotational_dynamics}, \eqref{eq_restoring_H}, and $d_{n,i}=s_i|d_{n,i}|$ gives
\begin{equation}
\begin{aligned}
\dot V={}&-c_{0,r}\omega\tanh\!\left(\frac{\omega}{\omega_c}\right)-c_{1,r}\omega^2\\
&+\sum_{i\in\mathcal A}|d_{n,i}|\omega s_i
\left(\left[\gamma_i-k_1f_{i0}\omega s_i\right]_+-\gamma_i\right).
\end{aligned}
\end{equation}
For each contact, if $\gamma_i-k_1f_{i0}\omega s_i\ge0$, then
\begin{equation}
\omega s_i\left(\left[\gamma_i-k_1f_{i0}\omega s_i\right]_+-\gamma_i\right)=-k_1f_{i0}\omega^2\le0.
\end{equation}

If $\gamma_i-k_1f_{i0}\omega s_i<0$, then
\begin{equation}
\omega s_i\left(\left[\gamma_i-k_1f_{i0}\omega s_i\right]_+-\gamma_i\right)=-\gamma_i\omega s_i.
\end{equation}
The first case is nonpositive. The second case requires $k_1>0$ and $\omega s_i>\gamma_i/(k_1f_{i0})>0$, so it is also negative. Thus, the controller contribution is nonpositive and is strictly dissipative for $\omega\neq0$ when $k_1>0$. Together with the rotational friction,
\begin{equation}
\dot V\le
-c_{0,r}\omega\tanh\!\left(\frac{\omega}{\omega_c}\right)
-c_{1,r}\omega^2
\le0.
\end{equation}
Under the conditions of the theorem, $\dot V=0$ only when $\omega=0$.

The barrier property in~\eqref{eq_potential_barrier} makes every sublevel set of $V$ compact and contained in $\Theta\times\mathbb R$. These sets are positively invariant because $\dot V\le0$. On $\omega=0$, \eqref{eq_projected_force_compact} gives $f_i=\gamma_i$. Substituting this into~\eqref{eq_rotational_dynamics} and using~\eqref{eq_restoring_H}, invariance requires
\begin{equation}
0=I_{\mathrm{cm}}\dot{\omega}
=\sum_{i\in\mathcal A}d_{n,i}\gamma_i(\theta)
=k_0H(\theta).
\end{equation}
By Lemma~1, this occurs only at $\theta=\theta^\star$. LaSalle's invariance principle therefore gives $(\theta(t),\omega(t))\to(\theta^\star,0)$.
\hfill$\square$

\subsection{Normal Force Resultant Alignment and Object Velocity}

The components of the normal force resultant along and perpendicular to the desired transport direction are
\begin{equation}
F_g=\sum_{i\in\mathcal A}f_i\cos\Delta_i,\qquad
F_p=\sum_{i\in\mathcal A}f_i\sin\Delta_i.
\end{equation}

\noindent\textbf{Corollary 1 (Asymptotic normal force resultant alignment).}
Under the conditions of Theorem~1 and Assumption~1,
\begin{equation}
\lim_{t\to\infty}F_p(t)=0,\qquad
\lim_{t\to\infty}F_g(t)=F_g^\star>0.
\end{equation}

\noindent\emph{Proof.}
Theorem~1 and the continuity of $[\cdot]_+$ give
\begin{equation}
f_i(t)\to\gamma_i^\star:=
\frac{k_0f_{i0}}{\sin|\Delta_i^\star|}>0,\qquad
\Delta_i^\star:=\theta^\star-a_i.
\end{equation}
Consequently,
\begin{equation}
F_p(t)\to
k_0\left(
\sum_{i\in\mathcal P}f_{i0}
-
\sum_{j\in\mathcal N}f_{j0}
\right)=0.
\end{equation}
The final equality follows from~\eqref{eq_group_balance}.

\begin{figure*}[!t]
\centering
\includegraphics[
    width=\textwidth,
    trim=0 0mm 0 0,
    clip
]{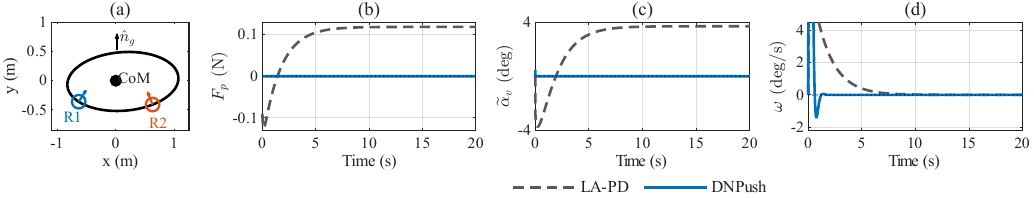}
\caption{Two-robot mechanism comparison under fixed nonparallel contacts. (a) Contact configuration. (b) Normal-force resultant component perpendicular to the desired transport direction. (c) Velocity direction error. (d) Object angular velocity.}
\label{fig_ellipse_ablation}
\vspace{-3mm}
\end{figure*}

Let $x_i^\star:=\theta^\star-a_i$ for $i\in\mathcal P$, $y_j^\star:=a_j-\theta^\star$ for $j\in\mathcal N$, and
$S:=\sum_{i\in\mathcal P}f_{i0}
=\sum_{j\in\mathcal N}f_{j0}>0$.
The limiting forward force satisfies
\begin{align}
\frac{F_g^\star}{k_0}
&=\sum_{i\in\mathcal P}f_{i0}\cot x_i^\star
+\sum_{j\in\mathcal N}f_{j0}\cot y_j^\star
\nonumber\\
&=\frac{1}{S}
\sum_{i\in\mathcal P}\sum_{j\in\mathcal N}
f_{i0}f_{j0}
\left(\cot x_i^\star+\cot y_j^\star\right)>0,
\end{align}
where the final inequality follows from~\eqref{eq_cotangent_pair_inequality}. Thus, $F_g^\star>0$.
\hfill$\square$

When $k_1=0$, Assumption~1 gives $F_p(t)=0$ throughout the nominal motion. When $k_1>0$, the angular rate term may produce a transient perpendicular force component, but this term vanishes as $\omega(t)\to0$.

\noindent\textbf{Theorem 2.}
Under the conditions of Theorem~1 and Assumption~1, consider the nominal translational model with zero tangential contact force resultant. If $c_{1,t}>0$, then
\begin{equation}
\bm v_G(t)\to v_g^\star\widehat{\bm n}_g,\qquad
v_g^\star>0.
\label{eq_velocity_component_limits}
\end{equation}
where $v_g^\star$ is the unique positive solution of
\begin{equation}
c_{0,t}\tanh\!\left(\frac{v_g^\star}{v_c}\right)
+c_{1,t}v_g^\star
=F_g^\star.
\label{eq_speed_equilibrium}
\end{equation}
Consequently, $v_p(t)\to0$, $v_g(t)\to v_g^\star$, and
$\widetilde{\alpha}_v(t)\to0$.

\noindent\emph{Proof.}
Corollary~1 gives
$\bm F_n(t)\to\bm F_n^\star:=F_g^\star\widehat{\bm n}_g$.
Define the translational resistance law
$\bm D_t(\bm v):=-\bm F_f(\bm v)$.
The left-hand side of~\eqref{eq_speed_equilibrium} is strictly increasing from zero to infinity, so $v_g^\star$ exists uniquely and
$\bm D_t(v_g^\star\widehat{\bm n}_g)=\bm F_n^\star$.

Let $\bm e:=\bm v_G-v_g^\star\widehat{\bm n}_g$. The smooth Coulomb component of $\bm D_t$ is the gradient of the convex function
$c_{0,t}v_c\log\cosh(\|\bm v\|/v_c)$, while its viscous component is strongly monotone. Thus,
\begin{equation}
\bm e^\top
\left[
\bm D_t(\bm v_G)
-
\bm D_t(v_g^\star\widehat{\bm n}_g)
\right]
\ge c_{1,t}\|\bm e\|^2.
\end{equation}

Using the translational dynamics and the bound above gives
\begin{equation}
\frac{m}{2}\frac{\mathrm d}{\mathrm dt}\|\bm e\|^2
\le
-\frac{c_{1,t}}{2}\|\bm e\|^2
+
\frac{\|\bm F_n-\bm F_n^\star\|^2}{2c_{1,t}}.
\end{equation}

Since $\bm F_n(t)-\bm F_n^\star\to\bm0$, a scalar comparison argument gives $\bm e(t)\to\bm0$, proving~\eqref{eq_velocity_component_limits}.
\hfill$\square$

\section{Results and Discussion}
We first evaluate the outer loop coordination mechanism in simulation with ideal force tracking, and then test the complete controller on differential-drive robots across different transport scenarios.

\subsection{Simulation Setup and Results}

All simulations use the same elliptical object with semiaxes $0.95$~m and $0.50$~m and its major axis oriented at $20^\circ$ in the body frame. The elliptical geometry provides a curved boundary, and the selected contact normals are nonparallel. Its geometric center is offset from the CoM by $[0.12,\ 0.02]^\top$~m in the body frame. The object parameters are $m=0.50$~kg and $I_{\mathrm{cm}}=0.10$~kg\,m$^2$. The resistance parameters are $c_{0,t}=c_{0,r}=0$, $c_{1,t}=1.0$~N\,s/m, and $c_{1,r}=1.0$~N\,m\,s/rad. Force commands are limited to $[0.5,5.0]$~N and $\varepsilon=10^{-4}$.

\begin{figure*}[t]
\centering
\includegraphics[
    width=\textwidth,
    trim=0mm 0 0mm 0,
    clip
]{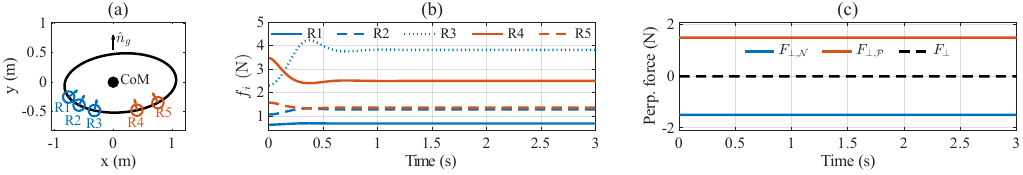}
\caption{Five-robot coordination with unequal moment sign groups. (a) Contact configuration with three robots in $\mathcal N$ and two in $\mathcal P$. (b) Individual normal force magnitudes. (c) Perpendicular force contributions $F_{p,\mathcal N}$ and $F_{p,\mathcal P}$ and their resultant $F_p$.}
\label{fig_five_robot}
\vspace{-2mm}
\end{figure*}
\subsubsection{Two-Robot Mechanism}

To isolate the reciprocal-sine force scaling mechanism, both controllers use the same two contacts, force limits, and nominal force scale, and differ only in how the normal force magnitude is determined. DNPush uses~\eqref{eq_force_policy} with $k_1=0$. The local-alignment PD (LA-PD) baseline instead uses proportional feedback on $\Delta_i$ and derivative feedback through $\omega=\dot{\Delta}_i$ to stabilize object rotation:
\begin{equation}
f_i^{\mathrm{LA}} = \operatorname{sat}\!\left[k_0f_{i0} -\left(K_\Delta\Delta_i +K_{\omega,\mathrm{LA}}\omega\right)
\operatorname{sgn}(d_{n,i})
\right].
\label{eq_sim_LAPD}
\end{equation}

The contacts are placed at ellipse parameter angles $t_1=215^\circ$ and $t_2=300^\circ$, giving $d_{n,1}=-0.341$~m and $d_{n,2}=0.437$~m, with $\Delta_1(0)=-32.49^\circ$ and
$\Delta_2(0)=21.34^\circ$. Both controllers use $k_0f_{i0}=1.5$~N. LA-PD uses $K_\Delta=1.0$~N/rad and $K_{\omega,\mathrm{LA}}=0.45$~N\,s/rad.

At a zero-angular rate equilibrium, the steady object velocity is aligned with the resultant of normal forces, so
\begin{equation}
\widetilde{\alpha}_v^\star
=
\operatorname{atan2}\!\left(
\sum_i f_i^\star\sin\Delta_i^\star,\,
\sum_i f_i^\star\cos\Delta_i^\star
\right).
\label{eq_sim_steady_direction}
\end{equation}
For the selected contacts and gains, solving the LA-PD zero-torque equilibrium predicts $\widetilde{\alpha}_v^\star=3.69^\circ$. Figure~\ref{fig_ellipse_ablation} confirms this behavior: LA-PD stabilizes the object rotation but retains a nonzero perpendicular force component and steady velocity direction error. DNPush also stabilizes the rotation while driving $F_p$ and the velocity direction error close to zero.
\subsubsection{Five-Robot Coordination}

We next evaluate DNPush with five contacts on the same ellipse at $t_i=\{200^\circ,220^\circ,240^\circ,285^\circ,310^\circ\}$, with signed moment arms approximately $\{-0.281,-0.333,-0.208,0.291,0.504\}$~m. Three contacts belong to $\mathcal N$ and two to $\mathcal P$. To keep the group weights consistent with the two-robot case, the nominal group weight of $1.5$~N is distributed equally within each group, giving $k_0f_{i0}=0.50$~N for $i\in\mathcal N$ and $0.75$~N for $i\in\mathcal P$. All robots use $k_1=0$, satisfying Assumption~1 with a total force weight of $1.5$~N per group.

The ordered contact condition remains satisfied throughout the simulation. As the object rotation converges, the individual normal forces converge to different steady values such that the net torque from the normal forces approaches zero, without using the moment arm magnitudes in the online controller. At steady state, the two groups generate equal and opposite perpendicular force contributions, with $F_{p,\mathcal N}\approx-1.50$~N and $F_{p,\mathcal P}\approx1.50$~N, so that $F_p\approx0$. The angular rate also converges to zero, and the steady velocity direction error is zero. Fig.~\ref{fig_five_robot} demonstrates coordinated force allocation using only local information and without inter-robot communication.

\subsubsection{Active Rotational Damping}

\begin{figure}[b]
\vspace{-5mm}
\centering
\includegraphics[width=0.95\linewidth]{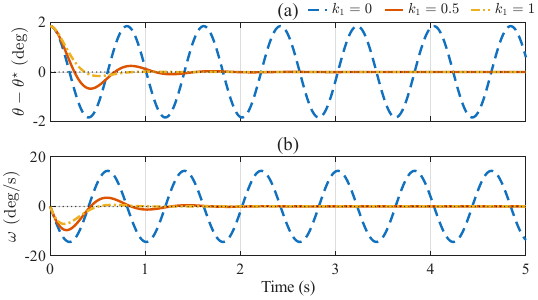}
\caption{Effect of active rotational damping with $c_{0,r}=c_{1,r}=0$. (a) Orientation error. (b) Object angular velocity.}
\label{fig_active_damping}
\end{figure}

To isolate the effect of the angular rate term, we repeat the two-robot simulation with $c_{0,r}=c_{1,r}=0$ while keeping the same contact geometry and force parameters. Fig.~\ref{fig_active_damping} compares $k_1=\{0,0.5,1.0\}$. With $k_1=0$, the restoring torque produces sustained rotational oscillation. For $k_1>0$, the oscillation decays, with the larger gain providing faster damping in this example.

\subsection{Hardware Experiments}

Hardware experiments use a rigid object and soft tactile differential-drive robots. The tactile sensor measures $F_{n,i}$ and $\phi_i^{(r)}$. The onboard controller was executed at approximately 10~Hz. We used OptiTrack to measure each robot's heading $\psi_i^{(w)}$ in the same world frame as the shared transport direction $\alpha^{(w)}$. The local DNPush law does not use another robot's pose, tactile data, or force command.

The transported object has a mass of approximately $0.90$~kg, and the effective Coulomb friction magnitude is $c_{0,t}\approx1.8$~N. Each robot uses the DNPush law with $k_1=0$ and $\varepsilon=0$. The tactile inner loop regulates the measured normal force using PI gains $K_{p,f}=40$ and $K_{i,f}=2$. The commanded normal force is limited to $0.5$--$4.5$~N. An additional two-robot waypoint trial is shown in the supplementary video.  \footnotemark 

\subsubsection{Fixed-Direction Repeatability with Two Robots}

\begin{figure}[t]
\vspace*{1.5mm}
\centering
\begin{subfigure}[t]{\linewidth}
\centering
\includegraphics[width=0.95\linewidth]{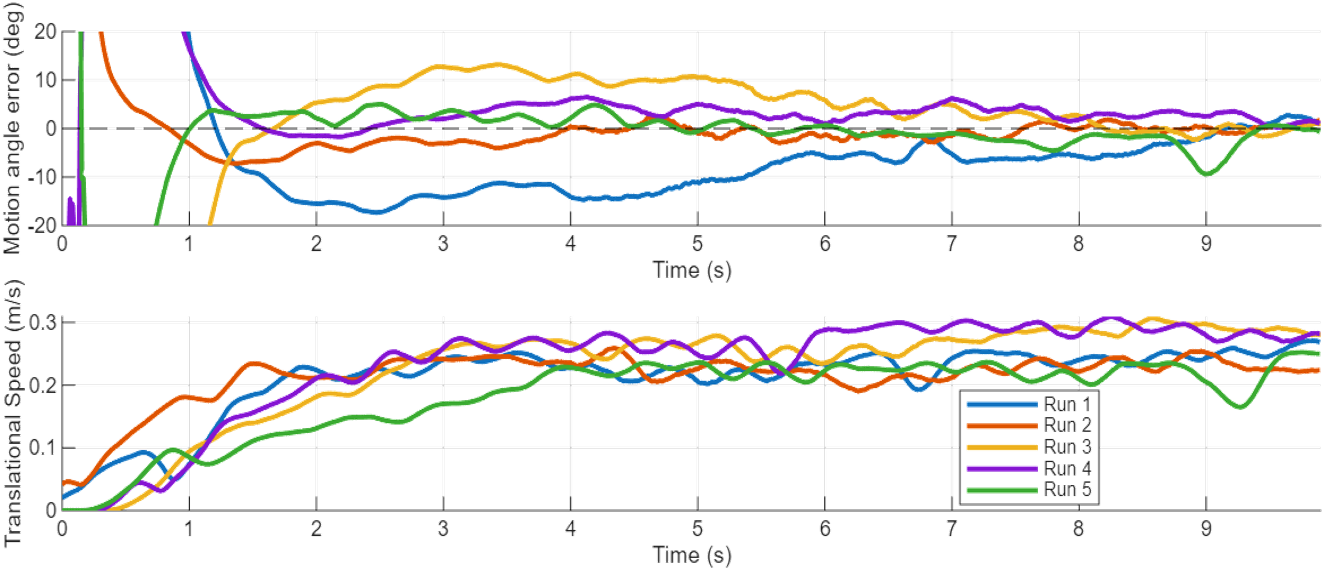}
\caption{Translational behavior over five trials.}
\end{subfigure}

\vspace{1mm}

\begin{subfigure}[t]{\linewidth}
\centering
\includegraphics[
    width=\linewidth,
    trim=0 0 0 3mm,
    clip
]{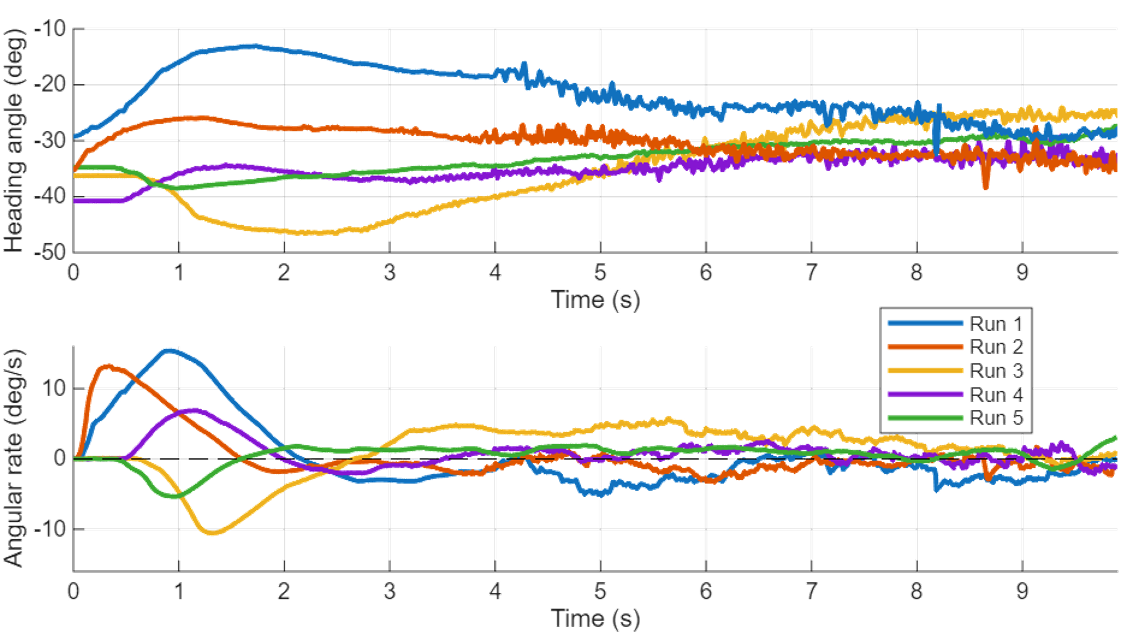}
\caption{Rotational behavior over five trials.}
\end{subfigure}

\caption{Fixed-direction hardware repeatability over five trials.}
\label{fig_hw_repeatability}
\vspace{-3mm}
\end{figure}

The fixed-direction study in Fig.~\ref{fig_hw_repeatability} evaluates repeatability under the same pushing contacts and desired transport direction. The two robots use equal reciprocal-sine force scales, $k_0f_{10}=k_0f_{20}=1.8$~N. Metrics are averaged over the final $0.5$~s of each of five trials and reported as mean and standard deviation. The velocity direction error across five trials is $0.61^\circ \pm 0.79^\circ$, the translational speed is $0.257 \pm 0.026$~m/s, and the angular rate is $0.31^\circ/\mathrm{s} \pm 0.90^\circ/\mathrm{s}$. The object heading settles to $-30.01^\circ \pm 3.54^\circ$, showing that the object velocity can align with the transport direction without requiring object heading alignment.

\subsubsection{Waypoint Tracking Using Three Robots}
\begin{figure}[t]
\vspace*{1.5mm}
\centering
\begin{subfigure}[t]{\linewidth}
\centering
\includegraphics[width=0.85\linewidth]{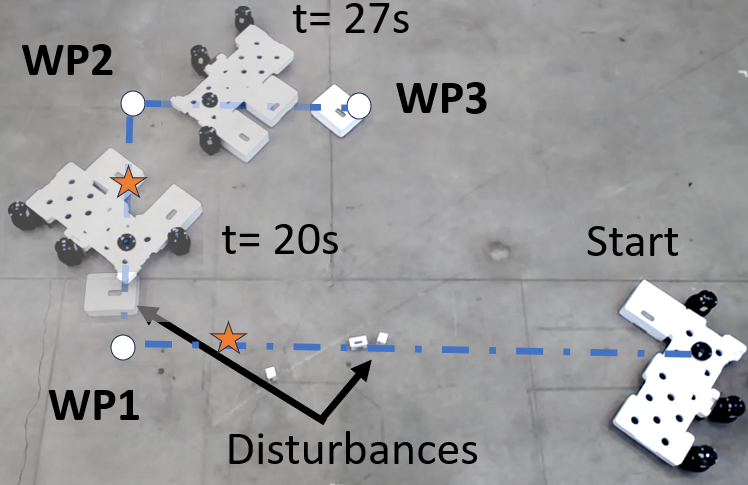}
\caption{Composite snapshots of the hardware trial.}
\end{subfigure}

\vspace{1mm}

\begin{subfigure}[t]{\linewidth}
\centering
\includegraphics[width=0.8\linewidth]{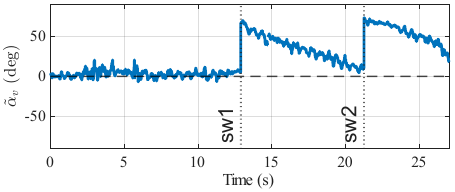}
\caption{Velocity direction error.}
\end{subfigure}

\caption{Three-robot waypoint tracking with disturbances.}
\label{fig_hw_three_robot}
\vspace{-4mm}

\end{figure}

Finally, we test waypoint tracking with three robots and small obstacles placed along the path as disturbances. At each control step, a fixed reference point on the object is used to define the shared desired transport direction toward the active waypoint, while the local DNPush force law does not use the object position. The target switches to the next waypoint when the reference point enters the waypoint tolerance region. Two contacts belong to $\mathcal N$ and one belongs to $\mathcal P$. The reciprocal-sine force scales are $k_0f_{10}=k_0f_{20}=1.1$~N for the two contacts in $\mathcal N$ and $k_0f_{30}=2.2$~N for the contact in $\mathcal P$, satisfying the group balance condition in~\eqref{eq_group_balance}. Fig.~\ref{fig_hw_three_robot} shows successful transport through the predefined waypoints using DNPush without inter-robot communication. The velocity direction error increases when the desired direction changes and then decreases as the object turns toward the new transport direction.
\subsection{Discussion and Limitations}

The analysis assumes fixed contacts, ideal force tracking, no upper force saturation, and $\varepsilon=0$. Rotational convergence further assumes zero tangential contact torque, while velocity convergence assumes zero tangential contact force resultant. The fixed-direction hardware trials directly evaluate the nominal transport objective, whereas the three-robot waypoint trial uses piecewise changes in the desired direction and is not covered by the fixed-direction convergence analysis. In hardware, finite force limits, tangential interactions, sensing noise, and tracking errors can produce residual errors. Fixed contact is not formally guaranteed, and future work will relax the fixed-contact assumption through online contact planning and contact-mode switching.
\footnotetext{Supplementary video: \href{https://www.youtube.com/watch?v=Q3_m8GYC6gM}{https://www.youtube.com/watch?v=Q3\_m8GYC6gM}}

\section{Conclusion}

This letter presented DNPush, a distributed pushing force controller for object transport without inter-robot communication. We showed that local force direction misalignment need not be eliminated in distributed pushing. Under an ordered opposite moment contact condition, the direction misalignment can instead be used as a local signal to balance the force components perpendicular to the desired direction and stabilize object rotation. Under fixed contacts and the nominal contact model, the analysis showed that object rotation asymptotically converges to a unique equilibrium, the resultant normal force aligns with the desired direction, and the object's velocity converges to the desired transport direction. Simulations isolated the force scaling mechanism, demonstrated five-robot coordination, and confirmed the effect of active rotational damping. Hardware experiments with soft tactile differential-drive robots achieve a velocity direction error of $0.61^\circ\pm0.79^\circ$ over five trials with a fixed object transport direction and further demonstrate three-robot object transport through predefined waypoints.

\bibliographystyle{IEEEtran}
\bibliography{References}

\end{document}